\pdfoutput=1

\documentclass[11pt]{article}

\usepackage[preprint]{acl}
\usepackage{times}
\usepackage{latexsym}
\usepackage[T1]{fontenc}
\usepackage[utf8]{inputenc}
\usepackage{microtype}
\usepackage{inconsolata}
\usepackage{graphicx}
\usepackage{tabularx} 
\usepackage{booktabs} 
\usepackage{array}    
\usepackage{booktabs}
\usepackage{multirow}
\usepackage{amsfonts}

\usepackage{graphicx}
\usepackage{amsmath}
\usepackage{amsthm}
\usepackage{booktabs}
\usepackage{algorithm}
\usepackage{algorithmic}
\usepackage[switch]{lineno}
\usepackage{xspace}
\usepackage{hyperref}

\usepackage{xcolor}
\usepackage{listings}
\usepackage{enumitem}
\setlist[enumerate]{  
    topsep=0pt,       
    itemsep=0pt,      
    parsep=2pt,       
    partopsep=0pt,    
    leftmargin=*      
}
\setlist[itemize]{    
    topsep=0pt,
    itemsep=0pt,
    parsep=2pt,
    partopsep=0pt,
    leftmargin=*
}

\usepackage[breakable,skins]{tcolorbox}

\newcommand{\makeideabox}[2]{  
    \begin{tcolorbox}[
      colback=blue!10,
      colframe=blue!70,
      arc=0pt,
      boxrule=1pt,
      title=#1,
      width=\textwidth,    
      breakable=true,      
      text width=\dimexpr\textwidth-2\tabcolsep\relax  
    ]
      #2
    \end{tcolorbox}
}

\definecolor{codegreen}{rgb}{0,0.6,0}
\definecolor{codegray}{rgb}{0.5,0.5,0.5}
\definecolor{codepurple}{rgb}{0.58,0,0.82}
\definecolor{backcolor}{rgb}{0.95,0.95,0.92}

\newcommand{\algname}{\textsc{FRAME}\xspace}

\title{FRAME: Feedback-Refined Agent Methodology for Enhancing Medical Research Insights}

\author{
 \textbf{Chengzhang Yu\textsuperscript{$~{1}$}}\thanks{Equal contribution},
 \textbf{Yiming Zhang\textsuperscript{${2,3*}$}},
 \textbf{Zhixin Liu\textsuperscript{$4$}},
\\
 \textbf{Zenghui Ding\textsuperscript{$2$}}\thanks{Correspondence to Zhanpeng Jin, Zenghui Ding: zjin@scut.edu.cn, dingzenghui@iim.ac.cn},
 \textbf{Yining Sun\textsuperscript{$2,3$}},
 \textbf{Zhanpeng Jin\textsuperscript{$1\dag $}}
\\
\\
 \textsuperscript{1}South China University of Technology,
\\
 \textsuperscript{2}HFIPS, Chinese Academy of Sciences,
\\
 \textsuperscript{3}University of Science and Technology of China,
\\
\textsuperscript{4}The Third Affiliated Hospital of SuYat-sen University
\\
}

\begin{document}
\maketitle
\begin{abstract}
The automation of scientific research through large language models (LLMs) presents significant opportunities but faces critical challenges in knowledge synthesis and quality assurance. We introduce Feedback-Refined Agent Methodology (\textbf{FRAME}), a novel framework that enhances medical paper generation through iterative refinement and structured feedback. Our approach comprises three key innovations: (1) A structured dataset construction method that decomposes 4,287 medical papers into essential research components through iterative refinement; (2) A tripartite architecture integrating Generator, Evaluator, and Reflector agents that progressively improve content quality through metric-driven feedback; and (3) A comprehensive evaluation framework that combines statistical metrics with human-grounded benchmarks. Experimental results demonstrate \textbf{FRAME}'s effectiveness, achieving significant improvements over conventional approaches across multiple models ($9.91\%$ average gain with DeepSeek V3, comparable improvements with GPT-4o Mini) and evaluation dimensions. Human evaluation confirms that \textbf{FRAME}-generated papers achieve quality comparable to human-authored works, with particular strength in synthesizing future research directions. The results demonstrated our work could efficiently assist medical research by building a robust foundation for automated medical research paper generation while maintaining rigorous academic standards.
\end{abstract}

\section{Introduction}\label{sec:Introduction}
The traditional academic research paradigm relies on human researchers to gather knowledge, formulate hypotheses, and evaluate findings through peer review. While this process has driven significant technological advances, it is inherently limited by human cognitive constraints and time-intensive workflows, with studies showing an average publication cycle of 21.9 months \cite{smartFactorsAssociatedConverting2013}. The emergence of large language models (LLMs), particularly since GPT-3.5 \cite{wuBriefOverviewChatGPT2023}, has introduced unprecedented capabilities in natural language processing, from text generation to complex reasoning tasks \cite{zhaoInvestigatingTabletoTextGeneration2023,wangRecentAdvancesInteractive2024,wangLLMbasedVisionLanguage2024,liElicitingTranslationAbility2024}. Advanced techniques like Chain-of-Thought (CoT) \cite{wei2022chain} and frameworks such as LangChain \cite{langchain_docs} and MetaGPT \cite{hong2024metagpt} have further enhanced LLMs' ability to handle sophisticated multi-agent collaboration.

\begin{figure*}[h]
    \centering
    \includegraphics[width=\textwidth]{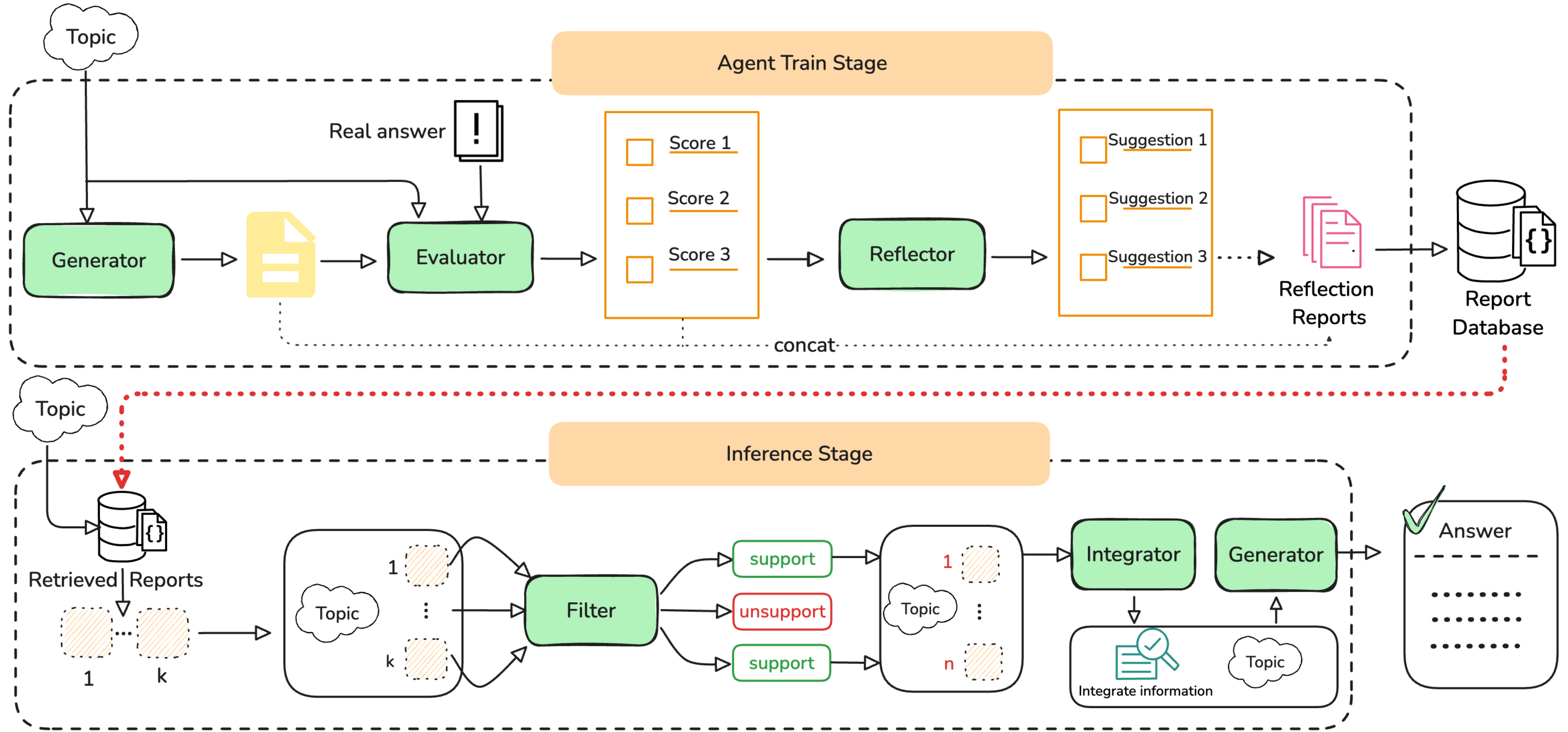}
    \caption{Architecture of the Feedback-Refined Agent Methodology (\textbf{FRAME}). During the training phase, the system generates and accumulates Reflection Reports in a dedicated database, which subsequently guides the formal paper generation process. This iterative training paradigm enables continuous refinement of the generation capabilities through structured feedback mechanisms.}
    \label{fig:training_process}
\end{figure*}

Current applications of LLMs in academic research fall into two categories. The first focuses on specific subtasks like code generation \cite{githubcopilot,cursor}, idea formulation \cite{huNovaIterativePlanning2024}, and paper review \cite{jinAgentReviewExploringPeer2024,sunMetaWriterExploringPotential2024}. While effective within their domains, these applications cannot encompass the entire research process. The second category employs collaborative multi-agent systems for comprehensive research tasks, exemplified by AutoSurvey's automated literature review pipeline \cite{wangAutoSurveyLargeLanguage2024a}. However, these applications have primarily focused on computational domains where experiments can be simulated.

The application of LLMs to support end-to-end academic research, particularly in medical research, faces two significant challenges. First, current large language models primarily rely on factual knowledge (e.g., understanding that colds can be caused by either bacteria or viruses) rather than learning from previous generation failures (e.g., recognizing that a previously unsuccessful attempt at generating a cold-related paper failed to adequately consider bacterial influences). This limitation hinders the models' ability to iteratively improve their output quality through experience-based learning \cite{wangAutoSurveyLargeLanguage2024a,luAIScientistFully2024}. Second, current paper generation models almost lack robust evaluation mechanisms, depending primarily on subjective agent assessments without rigorous benchmarks against human-authored papers. These limitations underscore the need for methods that ensure scientific validity while maintaining the academic standards characteristic of peer-reviewed publications.

Drawing inspiration from adversarial learning principles, we propose a novel approach to enhance the quality of LLM-generated academic papers, as shown in Figure \ref{fig:training_process}. Our Feedback-Refined Agent Methodology (\textbf{FRAME}) reimagines the dynamic interplay between content generation and quality assessment through a feedback-driven iterative process. Unlike traditional neural network approaches that rely on gradient-based parameter updates, \textbf{FRAME} implements a structured refinement cycle where specialized agents work in concert to progressively improve content quality. This methodology addresses the fundamental challenge of continuous self-improvement in LLM-based research systems through three key mechanisms: (1) Specialized agents assume distinct roles in content generation and quality evaluation; (2) The feedback process targets logical coherence and academic rigor through structured assessment; and (3) Knowledge accumulation is achieved through organized reflection reports that guide subsequent reasoning iterations.

Our contributions are summarized as follows:

\begin{enumerate}
\item We construct a comprehensive dataset of 4,287 medical research papers, covering diverse topics and methodologies, providing a robust knowledge base for training and evaluation.

\item We propose a Feedback-Refined Agent Methodology (\textbf{FRAME}) that effectively leverages prior research, demonstrating superior performance in generating high-quality medical research papers.
  
\item We introduce a novel evaluation method using human-authored papers as the gold standard, enabling objective assessment of generated content quality in medical research.
\end{enumerate}

\section{Related Works}\label{sec:Releated}
\subsection{LLMs for applications in the medical field}
The integration of LLMs into medical practice has demonstrated transformative potential across documentation, education, and diagnostics while facing persistent challenges in reliability and ethical governance. In medical writing and research management, models like ChatGPT streamline manuscript drafting, literature synthesis, and multidisciplinary data integration—exemplified by applications in ophthalmology for surgical summaries and cross-disciplinary research coordination \cite{peng2023study,bernstein2023comparison,gu2023plan}. However, limitations, including occasional reference fabrication and superficial contextual understanding, necessitate robust fact-checking mechanisms and integration with validated medical databases \cite{nakaura2023writing,thapa2023chatgpt}. Educational applications leverage LLMs for simulating patient interactions, generating practice questions, and automating assessments \cite{dave2023chatgpt,cascella2023evaluating}, while diagnostic implementations assist in parsing unstructured clinical data and improving doctor-patient communication \cite{benary2023leveraging,dias2019artificial}. Despite these advancements, critical gaps persist in causal reasoning and contextual interpretation, particularly for ambiguous clinical scenarios \cite{harris2023large}. The collective experience underscores the necessity for rigorous validation protocols, cultural contextualization, and ethical frameworks to ensure LLMs augment rather than compromise medical standards, serving as decision-support tools rather than autonomous agents \cite{shah2023creation,peng2023study}.

\subsection{AI-Powered Academic Research Methods}

The evolution of AI in scientific paper writing has progressed from rudimentary language correction to sophisticated end-to-end manuscript generation. Early applications focused on foundational tasks such as grammar and domain-specific spelling checks, with specialized models trained to identify errors in technical fields like medical academic texts \cite{lai2015automated}. As AI advanced, its role expanded to address broader linguistic challenges, particularly for non-native English speakers in the English-as-a-Foreign-Language (EFL) community. Tools like Wordtune emerged, enabling writers to overcome language barriers by dynamically rephrasing content across tones (e.g., formal vs. casual) and lengths (e.g., concise summaries or expanded explanations), thereby enhancing both clarity and stylistic adaptability \cite{zhao2023leveraging}. The advent of large language models (LLMs) marked a paradigm shift, empowering AI systems to automate complex scholarly workflows. For instance, AutoSurvey systematizes literature review creation in fast-evolving domains (e.g., AI research) through a structured pipeline: preliminary data retrieval and outline generation, subsection drafting via specialized LLMs, content integration, and iterative refinement \cite{wangAutoSurveyLargeLanguage2024a}. Further pushing boundaries, AI-Scientist demonstrates end-to-end research automation—generating hypotheses, designing experiments, analyzing results, and drafting full manuscripts—while even simulating peer-review processes to evaluate scientific rigor \cite{luAIScientistFully2024}. These advancements underscore AI's growing capacity to augment—though not yet fully replace—human ingenuity in academic writing, balancing efficiency gains with persistent demands for domain expertise and critical oversight.

\section{Dataset Construction}
\subsection{Existing Dataset Challenges}

\begin{figure*}[h]
    \centering
    \includegraphics[width= \textwidth]{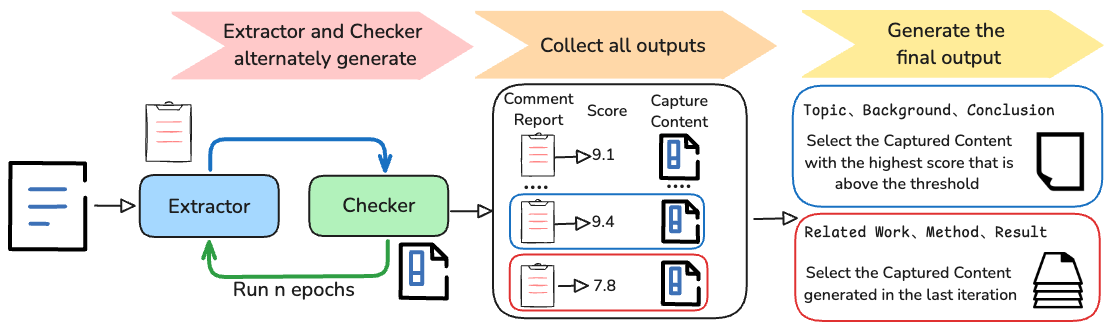}
    \caption{Overview of the dataset construction process. Core information from academic papers is iteratively extracted and refined through $N$ rounds of \emph{Extractor}-\emph{Checker} cycles ($N=3$ in our implementation), resulting in a more structured and concise representation of the content.}
    \label{fig:data_extract}
\end{figure*}

The construction of datasets for automated paper generation has long followed a paradigm focused on content aggregation, where researchers primarily harvest publicly available academic papers to build monolithic document repositories. For instance, AutoSurvey \cite{wangAutoSurveyLargeLanguage2024a} simply vectorized and stored $530,000$ arXiv computer science papers as its retrieval database without sophisticated preprocessing. Similarly, AI-Scientist\cite{luAIScientistFully2024} utilized $500$ papers from ICLR 2022 to evaluate the reliability of its proposed paper assessment model. Another example is Nova \cite{huNovaIterativePlanning2024}, which leveraged a total of 170 papers from CVPR 2024, ACL 2024, and ICLR 2024 to generate initial idea seeds.

However, these existing approaches exhibit two significant limitations in their utilization of academic papers. First, they primarily treat the papers as a static knowledge repository, failing to deeply extract and analyze the structural and logical frameworks inherent in the papers. Second, the overall processing pipeline is often overly simplistic, lacking tailored extraction methods for different sections of the papers. These limitations hinder the ability to fully leverage the rich information embedded in academic papers, thereby constraining the potential of automated paper generation systems.

To address these shortcomings, we propose a more refined approach to dataset construction. Specifically, we decompose each paper into six distinct sections: Topic, Background, Related Work, Method, Result, and Conclusion. The Topic section captures the specific research question or problem the paper aims to address. The Background section provides the contextual foundation and motivation for the study. The Related Work section identifies and discusses the connections between the paper and prior research. The Method section details the research methodology employed, while the Result section presents experimental data, including specific numerical findings. Finally, the Conclusion section summarizes the key takeaways and implications of the study. By extracting and analyzing these sections individually, we aim to create a more robust and granular dataset that can serve as a stronger foundation for subsequent paper generation tasks. This approach not only enhances the depth of information extraction but also enables more targeted and context-aware generation processes.

\subsection{Dataset Construction Process}

Our dataset construction begins with an initial corpus of $10{,}000$ research articles harvested from the medRxiv platform, covering $51$ medical disciplines (e.g., Allergy and Immunology, HIV/AIDS, Emergency Medicine) and published between 2023 and 2024. To ensure the quality and relevance of the data, we employed a two-tier selection protocol: (1) preference was given to articles subsequently accepted by peer-reviewed journals or cited in other scholarly works; (2) for the remaining articles, we utilized LLM to automatically screen for methodological rigor and completeness, discarding submissions that failed to meet these criteria.

We then implemented a systematic \textit{chapter-to-section mapping} protocol to align the structural components of the research articles with our dataset schema. This mapping ensures that each article adheres to conventional academic organization, thereby validating structural integrity while standardizing content extraction. Specifically, key components are mapped as follows:

\begin{itemize}
    \item \textbf{Topic (Top.)}:Focused on the research question or problem statement.
    \item \textbf{Background (Bkgd.)}: Derived from introductory material establishing the study context and motivation.
    \item \textbf{Related Work (RelW.)}: Sourced directly from dedicated literature review sections.
    \item \textbf{Method (Meth.)}: Extracted from the methodology sections detailing research designs and procedures.
    \item \textbf{Result (Res.)}: Captured from empirical data and analyses presented in the results section.
    \item \textbf{Conclusion (Conc.)}: Synthesized from discussions of findings and implications in the conclusion.
\end{itemize}

To further ensure data quality, we employed a three-stage filtration process. First, the initial selection criteria (journal acceptance, citation status, and LLM screening) eliminated fundamentally flawed submissions. Second, we excluded articles with non-standard section titles that did not align with our predefined aliases (see Table \ref{tab:section_table}), effectively removing those lacking essential structural components. Finally, during the agent-based extraction process (detailed in Section \ref{Data Extraction Agent}), we discarded samples with failed extraction attempts. This multi-stage filtering progressively enforced content quality, structural completeness, and technical consistency, resulting in a final curated dataset of $4{,}287$ high-quality medical research articles.

\subsection{Data Extraction Agent}\label{Data Extraction Agent}
To ensure the accuracy and reliability of information extraction from academic papers, we design a dual-agent framework consisting of an \emph{Extractor} and a \emph{Checker} for each section of the paper. As shown in Figure \ref{fig:data_extract}, this iterative refinement process enables progressive quality improvement through structured feedback loops.

The workflow operates through three key phases:
\begin{enumerate}
    \item The \emph{Extractor} first retrieves key information from designated chapters using the mapping defined in Table \ref{tab:section_table}, yielding the Captured Content that will serve as the foundation for all subsequent processing steps.
    \item The \emph{Checker} then evaluates the Captured Content using the evaluation metrics specified in Table \ref{tab:section_table}, assigning quantitative scores (1-5 scale) for each metric
    \item Based on the evaluation results, the \emph{Checker} generates targeted improvement suggestions that are fed back to the \emph{Extractor} for the next iteration
\end{enumerate}

\begin{table}[h]
    \centering
    \begin{tabularx}{\columnwidth}{>{\centering\arraybackslash}p{0.4cm}>{\centering\arraybackslash}X>{\centering\arraybackslash}X}
        \toprule
        \textbf{Section} & \textbf{Capture} & \textbf{Candidate Alias} \\
        \midrule
        Top. & Introduction & introduction \\
        \addlinespace
        Bkgd. & Introduction & introduction \\
        \addlinespace
        RelW. & Related Work & related work, literature review, previous work \\
        \addlinespace
        Meth. & Methodology & methods, methodology \\
        \addlinespace
        Res. & Experiment & results, findings, outcomes \\
        \addlinespace
        Conc. & Conclusion or Discussion & conclusions, summary, discussion \\
        \bottomrule
    \end{tabularx}
    \caption{Mapping Between Sections and Captured Chapters with Candidate Aliases, indicating acceptable chapter titles for content extraction}
    \label{tab:section_table}
\end{table}

This cyclic process continues for $n$ rounds, with each iteration producing progressively refined outputs. We implement different strategies for different sections based on their typical complexity and length characteristics.

For the Related Work, Method, and Result sections, which are typically longer and more structured, we adopt a progressive refinement approach. The output from the final iteration of the Extractor is selected by default.

\begin{table}[!t]
    \centering
    \begin{tabularx}{\columnwidth}{>{\centering\arraybackslash}p{0.6cm}>{\centering\arraybackslash}X>{\centering\arraybackslash}X}
        \toprule
        \textbf{Section} & \textbf{Evaluation Metric} & \textbf{Recommendation Dimensions} \\
        \midrule
        Bkgd. & Completeness, Relevance, Organization & Content, Relevance, Structure, Style \\
        \addlinespace
        RelW. & Diversity, Criticality, Connection & Coverage, Analysis, Connection, Synthesis \\
        \addlinespace
        Meth. & Coherence, Necessity, Completeness & Completeness, Flow, Precision, Justification \\
        \addlinespace
        Conc. & Comprehensiveness, Impact, Future Direction & Summary, Impact, Future, Synthesis \\
        \bottomrule
    \end{tabularx}
    \caption{Section-Specific Evaluation Metrics and Corresponding Recommendation Dimensions}
    \label{tab:evaluation_table}
\end{table}

For the typically shorter but often conceptually dense Topic, Background, and Conclusion sections, we implement a quality-gated selection mechanism. Due to their higher complexity-to-length ratio, the refinement process may not always be monotonic. Therefore, only iterations exceeding predefined quality thresholds across all relevant metrics are considered, and the highest-scoring iteration among those that meet the threshold requirements is selected. If none of the iterations surpass the threshold, the extraction is deemed unsuccessful.

To further enhance the extraction accuracy for the Result section, we employ a preprocessing step using regular expressions to identify and extract all numerical values from the input text. Irrelevant numbers (e.g., page numbers, section numbers) are filtered out through pattern matching and contextual analysis, with the remaining numerical data provided as additional context to the \emph{Extractor}. This focused approach enables the system to prioritize experimentally significant results while maintaining methodological rigor.

\section{Feedback-Refined Agent Methodology (FRAME)}
To enable our Agent to craft high-quality medical papers on par with human standards, it is insufficient to merely furnish the Agent with a wealth of background knowledge. Instead, the Agent must be capable of explicitly learning from each deficiency and updating its generation strategy in subsequent iterations. Drawing inspiration from adversarial learning principles, we have developed a feedback-driven iterative methodology known as the Feedback-Refined Agent Methodology (\textbf{FRAME}), whose architecture is illustrated in Figure \ref{fig:training_process}. This approach emphasizes continuous improvement through structured feedback loops, where specialized agents work in concert to progressively enhance content quality while maintaining rigorous academic standards.

\subsection{Agent Training Stage}
Our training framework employs a tripartite architecture that synergistically combines generation, evaluation, and reflection mechanisms. As depicted in Figure \ref{fig:training_process}, the system operates through three core components:

\begin{enumerate}
    \item The \emph{Generator} synthesizes manuscript sections (e.g., Background, Related Work, Method) conditioned on the research topic
    \item The \emph{Evaluator} conducts multi-dimensional quality assessments using the evaluation metrics defined in Table \ref{tab:evaluation_table} (Column 2), producing quantitative scores (1-5 scale) for each metric
    \item The \emph{Reflector} translates \emph{Evaluator} feedback into structured reflection reports by mapping criticism to specific suggestion dimensions from Table \ref{tab:evaluation_table} (Column 3)
\end{enumerate}

This triadic interaction creates a closed-loop learning system where each component informs the others' improvements. The Evaluator's metric-driven scoring (e.g., assessing Method section coherence through the "Coherence, Necessity, Completeness" metrics) provides objective performance measures, while the \emph{Reflector}'s dimension-specific recommendations (e.g., "Improve Flow and Justification" for Methods) guide targeted revisions.

\subsection{Inference Stage}
In the phase of generating new papers, our objective is to ensure that the Agent effectively leverages the valuable insights gained during the training phase while disregarding irrelevant experiences. Moreover, it is crucial to manage the context length to optimize the model's generation performance.

We employ a \emph{Retrieval-Augmented Generation (RAG)} approach to retrieve N Reflection Reports from the database, which were formulated during the training phase. This method focuses on extracting substantive insights rather than merely related information, ensuring a stronger foundation for the generation process.

Subsequently, we introduce a model known as the \emph{Filter}, which acts as a gatekeeper by eliminating reports that appear proximal in the vector space but are not truly relevant. This filtering step effectively reduces the interference of unrelated experiences, allowing the Agent to focus on generating content that is more aligned with the intended objectives.

Directly inputting multiple reports into the \emph{Generator} may result in excessive context length, which could impair the Agent's ability to focus on crucial information. Conversely, relying solely on a single report might lead to incomplete understanding, as individual reports typically contain only partial experiential insights. To address this challenge, we utilize an \emph{Integrator} to consolidate and merge multiple pertinent reports. This process is akin to using a larger batch size in Neural Network rather than a batch size of one, facilitating the acquisition of more balanced and comprehensive information. This approach helps the Agent avoid misleading influence from extreme data points.

\section{Experiment}\label{sec:Experiment}

Our experimental evaluation comprises four key components. First, we systematically evaluate the impact of \algname on text generation quality by comparing outputs from two state-of-the-art language models (DeepSeek V3 and GPT-4o Mini) with and without our method. Second, we conduct a rigorous human evaluation where $20$ randomly selected \algname-enhanced papers generated by DeepSeek V3 are compared against human-authored counterparts through expert assessments by medical professionals. Third, we investigate the generalizability of \algname by evaluating its performance on models released prior to our test set creation, ensuring no potential data contamination. Finally, we examine the influence of training dataset scale on model performance by conducting systematic experiments with varying dataset sizes, providing insights into the data efficiency of our approach. 

\subsection{Experimental Setup}

All experiments were conducted on an A800 computer to ensure a consistent and controlled environment for evaluating model performance across various tasks and metrics. The models DeepSeek v3 and GPT-4o Mini were accessed via their official APIs, while other models were deployed using PyTorch and vLLM \cite{kwon2023efficient}. The dataset, comprising a total of $4,287$ samples, was carefully partitioned to avoid temporal data leakage. Specifically, we used September 1, 2024, as the cutoff point to divide the dataset into training and testing sets, ensuring no temporal overlap between the two. The training set consists of $4,119$ samples, while the testing set contains $168$ samples. We employed FAISS as our database software, which allows for flexible data management, including the ability to freely add or remove entries, facilitating incremental training in future updates.
\subsection{Evaluation Metrics}
In this study, we employ two distinct sets of evaluation metrics to assess the performance of our models. The first set comprises statistical metrics, specifically Soft Precision and Soft Recall, which measure the alignment between the model's output and the ground truth. Soft Precision is defined as the ratio of correctly predicted relevant elements to the total number of predicted relevant elements, while Soft Recall is the ratio of correctly predicted relevant elements to the total number of actual relevant elements. These metrics provide a nuanced understanding of the model's accuracy and coverage, particularly in scenarios where binary classification is insufficient.

$$
\text{Prec.} = \frac{\sum_{i=1}^{n} \text{Sim}(P_i, G_i) \cdot \mathbb{I}(P_i \in \text{Relevant})}{\sum_{i=1}^{n} \mathbb{I}(G_i \in \text{Relevant})},
$$

$$
\text{Rec.} = \frac{\sum_{i=1}^{n} \text{Sim}(P_i, G_i) \cdot \mathbb{I}(G_i \in \text{Relevant})}{\sum_{i=1}^{n} \mathbb{I}(G_i \in \text{Relevant})},
$$

where $P_i$ represents the $i$-th predicted element, $G_i$ denotes the corresponding ground truth element, and $\text{Sim}(P_i, G_i)$ quantifies the similarity between $P_i$ and $G_i$. The function $\mathbb{I}(P_i \in \text{Relevant})$ is an indicator that returns 1 if $P_i$ is deemed relevant and 0 otherwise, while $\mathbb{I}(G_i \in \text{Relevant})$ indicates whether the $i$-th ground truth element is relevant.

The second set of metrics involves LLM-based scoring, where the output is evaluated across multiple dimensions such as Background, Related Work, Method, and Conclusion. For each dimension, the model's output is compared against the topic and the ground truth, and a score ranging from 1 to 5, in increments of 0.1, is assigned. To mitigate the randomness in LLM-based evaluations, we conduct three independent assessments for the same content and dimension, and the final score is computed as the average of the three evaluations. This scoring system allows for a detailed and robust assessment of the model's performance in generating coherent and contextually relevant content. The specific evaluation dimensions are detailed in Table \ref{tab:evaluation_table}.
\subsection{Model Comparisons}

\begin{table}[!t]
\centering
\footnotesize
\setlength{\tabcolsep}{0.3em}
\renewcommand{\arraystretch}{0.95}
\begin{tabular}{@{}lcclllll@{}}
\toprule
\multirow{2}{*}{Sect.} & \multirow{2}{*}{Method} & \multicolumn{2}{c}{Metrics (\%)} & \multicolumn{3}{c}{LLM Scores} \\
\cmidrule(lr){3-4} \cmidrule(lr){5-7}
 & & Prec. & Rec. & S1 & S2 & S3 & Total \\
\midrule
\multirow{4}{*}{Bkgd.} 
 & \textbf{Ours} & \textbf{90.98} & \textbf{89.50} & \textbf{74.67} & \textbf{83.67} & \textbf{82.82} & \textbf{84.33} \\
 & Filter & 87.33 & 86.28 & 59.01 & 65.30 & 64.78 & 72.54 \\
 & RAG & 87.12 & 85.85 & 57.42 & 63.39 & 62.70 & 71.30 \\
 & No-RAG & 87.64 & 86.15 & 55.94 & 61.91 & 63.24 & 70.98 \\

\multirow{4}{*}{RelW.}
 & \textbf{Ours} & \textbf{100.0} & \textbf{98.46} & \textbf{86.98} & \textbf{90.90} & \textbf{92.97} & \textbf{93.86} \\
 & Filter & 99.10 & 97.36 & 83.04 & 86.39 & 88.93 & 90.96 \\
 & RAG & 95.95 & 94.49 & 73.94 & 77.26 & 78.77 & 84.08 \\
 & No-RAG & 95.32 & 94.22 & 74.15 & 77.74 & 79.38 & 84.16 \\

\multirow{4}{*}{Meth.}
 & \textbf{Ours} & \textbf{98.87} & \textbf{94.08} & \textbf{83.29} & \textbf{85.96} & \textbf{78.96} & \textbf{88.23} \\
 & Filter & 98.68 & 92.41 & 79.54 & 83.12 & 73.82 & 85.52 \\
 & RAG & 94.58 & 89.96 & 75.89 & 80.11 & 70.98 & 82.31 \\
 & No-RAG & 95.50 & 91.52 & 79.05 & 82.59 & 74.62 & 84.66 \\

\multirow{4}{*}{Conc.}
 & \textbf{Ours} & \textbf{93.74} & \textbf{93.74} & \textbf{73.88} & \textbf{76.59} & \textbf{77.03} & \textbf{83.00} \\
 & Filter & 92.98 & 92.98 & 69.74 & 71.44 & 72.73 & 79.97 \\
 & RAG & 89.79 & 89.79 & 55.94 & 57.30 & 58.30 & 70.22 \\
 & No-RAG & 90.07 & 90.07 & 58.54 & 60.14 & 60.29 & 71.82 \\
\bottomrule
\end{tabular}
\caption{Performance comparison of DeepSeek V3 with different methods. S1-S3 represent distinct evaluation dimensions across different sections (e.g., background, methodology), with detailed descriptions provided in Table \ref{tab:evaluation_table}.}
\label{tab:method_comparison_deepseek}
\end{table}

\begin{table}[!t]
\centering
\footnotesize
\setlength{\tabcolsep}{0.3em}
\renewcommand{\arraystretch}{0.95}
\begin{tabular}{@{}lcclllll@{}}
\toprule
\multirow{2}{*}{Sect.} & \multirow{2}{*}{Method} & \multicolumn{2}{c}{Metrics (\%)} & \multicolumn{3}{c}{LLM Scores} \\
\cmidrule(lr){3-4} \cmidrule(lr){5-7}
 & & Prec. & Rec. & S1 & S2 & S3 & Total \\
\midrule
\multirow{4}{*}{Bkgd.} 
 & \textbf{Ours} & \textbf{90.48} & \textbf{90.07} & \textbf{73.44} & \textbf{82.08} & \textbf{80.07} & \textbf{83.23} \\
 & Filter & 89.24 & 88.34 & 65.35 & 72.86 & 72.26 & 77.61 \\
 & RAG & 89.21 & 88.33 & 66.03 & 73.37 & 72.65 & 77.92 \\
 & No-RAG & 89.37 & 88.52 & 65.66 & 72.92 & 72.83 & 77.86 \\

\multirow{4}{*}{RelW.}
 & \textbf{Ours} & \textbf{99.77} & \textbf{97.44} & \textbf{80.79} & \textbf{84.97} & \textbf{87.96} & \textbf{90.19} \\
 & Filter & 97.17 & 95.08 & 74.36 & 78.86 & 81.29 & 85.35 \\
 & RAG & 97.20 & 95.07 & 74.06 & 78.43 & 81.15 & 85.18 \\
 & No-RAG & 97.11 & 95.12 & 74.05 & 78.59 & 81.17 & 85.21 \\

\multirow{4}{*}{Meth.}
 & \textbf{Ours} & \textbf{98.78} & \textbf{91.50} & \textbf{77.43} & \textbf{81.37} & \textbf{71.71} & \textbf{84.16} \\
 & Filter & 97.05 & 89.86 & 74.90 & 79.38 & 67.86 & 81.81 \\
 & RAG & 96.51 & 89.57 & 74.73 & 79.04 & 68.33 & 81.64 \\
 & No-RAG & 96.52 & 89.64 & 74.86 & 79.19 & 68.04 & 81.65 \\

\multirow{4}{*}{Conc.}
 & \textbf{Ours} & \textbf{93.27} & \textbf{93.27} & \textbf{74.23} & \textbf{77.27} & \textbf{78.25} & \textbf{83.26} \\
 & Filter & 91.77 & 91.77 & 68.37 & 71.40 & 72.38 & 79.14 \\
 & RAG & 91.69 & 91.69 & 67.64 & 70.36 & 71.47 & 78.57 \\
 & No-RAG & 92.24 & 92.24 & 70.90 & 73.81 & 74.89 & 80.81 \\
\bottomrule
\end{tabular}
\caption{Performance comparison of GPT-4o Mini with different methods. S1-S3 represent distinct evaluation dimensions across different sections (e.g., background, methodology), with detailed descriptions provided in Table \ref{tab:evaluation_table}.}
\label{tab:method_comparison_gpt}
\end{table}

As demonstrated in Tables \ref{tab:method_comparison_deepseek} and \ref{tab:method_comparison_gpt}, our method exhibits consistent superiority across both mainstream models. The comprehensive evaluation encompassing $40$ comparative dimensions ($4$ sections × [5 metrics (soft precision, soft recall, S1, S2, S3)] × $2$ models) reveals two key findings when compared against three clearly defined baselines: (1) No-RAG, which uses direct model inference without retrieval augmentation; (2) standard RAG, which retrieves relevant content from a FAISS vector database and concatenates it with model input; and (3) Filter, an ablation of our GIA framework that evaluates retrieved content using an LLM. Our approach outperforms all these baselines in every experimental configuration, with an average performance improvement of $9.91\%$ across all sections compared to the strongest baseline (calculated as the mean of five evaluation metrics per section). These results substantiate the effectiveness of our method in enhancing scientific paper generation across different model architectures and document sections.

\subsection{LLM Knowledge Cutoff and Paper Generation}

\begin{table}[!t]
  \centering
  \footnotesize
  \setlength{\tabcolsep}{0.3em}
  \renewcommand{\arraystretch}{0.95}
  \begin{tabular}{@{}lcclllll@{}}
  \toprule
  \multirow{2}{*}{Sect.} & \multirow{2}{*}{Method} & \multicolumn{2}{c}{Metrics (\%)} & \multicolumn{3}{c}{LLM Scores} \\
  \cmidrule(lr){3-4} \cmidrule(lr){5-7}
   & & Prec. & Rec. & S1 & S2 & S3 & Total \\
  \midrule
  \multirow{4}{*}{Bkgd.}
   & \textbf{Ours} & \textbf{89.70} & \textbf{88.37} & \textbf{72.07} & \textbf{78.37} & \textbf{76.25} & \textbf{80.95} \\
   & Filter & 87.72 & 87.61 & 66.48 & 71.90 & 68.79 & 76.50 \\
   & RAG & 85.70 & 87.52 & 63.58 & 64.76 & 60.86 & 72.48 \\
   & No-RAG & 89.05 & 87.49 & 68.51 & 76.23 & 75.80 & 79.41 \\

  \multirow{4}{*}{RelW.}
   & \textbf{Ours} & \textbf{97.41} & \textbf{95.91} & \textbf{76.46} & \textbf{77.91} & \textbf{82.31} & \textbf{86.00} \\
   & Filter & 96.08 & 94.77 & 75.69 & 77.75 & 82.18 & 85.30 \\
   & RAG & 95.65 & 94.08 & 70.63 & 73.90 & 77.94 & 82.44 \\
   & No-RAG & 95.06 & 94.13 & 74.00 & 76.58 & 80.02 & 83.96 \\

  \multirow{4}{*}{Meth.}
   & \textbf{Ours} & \textbf{97.16} & \textbf{90.24} & \textbf{77.64} & \textbf{81.34} & \textbf{68.71} & \textbf{83.02} \\
   & Filter & 96.44 & 90.07 & 74.24 & 78.90 & 67.59 & 81.45 \\
   & RAG & 96.82 & 90.09 & 75.07 & 79.00 & 68.34 & 81.87 \\
   & No-RAG & 96.58 & 89.84 & 75.73 & 80.24 & 68.53 & 82.19 \\

  \multirow{4}{*}{Conc.}
   & \textbf{Ours} & \textbf{92.96} & \textbf{92.96} & \textbf{74.36} & \textbf{72.54} & \textbf{70.83} & \textbf{80.73} \\
   & Filter & 92.10 & 92.10 & 73.30 & 76.82 & 76.74 & 82.21 \\
   & RAG & 91.99 & 91.99 & 72.13 & 75.16 & 75.49 & 81.35 \\
   & No-RAG & 91.95 & 91.95 & 70.72 & 72.61 & 68.15 & 79.08 \\
  \bottomrule
  \end{tabular}
  \caption{Performance comparison of Qwen1.5-32B with different methods. S1-S3 represent distinct evaluation dimensions across different sections (e.g., background, methodology), with detailed descriptions provided in Table
  \ref{tab:evaluation_table}.}
  \label{tab:method_comparison_qwen}
  \end{table}

To validate the effectiveness of our proposed \algname method across smaller-scale language models and to mitigate potential biases arising from the inclusion of this paper in LLM pretraining datasets, we conducted experiments using Qwen 1.5 32B (released on February 6, 2024) on a test set comprising papers published on or after September 1, 2024. The experimental results demonstrate that Qwen 1.5 32B achieves comparable performance to DeepSeek V3 without our method, indicating that long-chain paper reasoning does not significantly differ between non-reasoning-focused models. However, as demonstrated in Table \ref{tab:method_comparison_qwen}, the implementation of \algname yields a substantial performance improvement of $1.5\%$ ($81.16\%$ vs $82.68\%$), thereby substantiating the efficacy of our approach in enhancing the quality of generated scientific papers across different model architectures and scales.

\subsection{Human Assessment}

To objectively evaluate the effectiveness of \algname, we conducted a human assessment study comparing 20 papers generated by \algname-enhanced DeepSeek V3 with 20 human-authored papers. A panel of medical professionals evaluated both sets of papers across key sections including Background, Methodology, Results, and Conclusion.

As shown in Figure \ref{fig:human_model}, the evaluation results demonstrate that \algname-generated papers achieve statistically equivalent quality to human-authored papers across most sections, with no significant difference in composite scores ($M_{\text{model}} = 92.80\%$ vs $M_{\text{human}} = 92.88\%$, $p = 0.746$, Cohen's $d = -0.098$). However, in synthesizing future research directions—a critical aspect of scientific writing—\algname exhibits superior performance compared to human authors ($p < 0.001$, $d = 2.27$), suggesting particular effectiveness in forward-looking synthesis.
\begin{figure}[!htbp]
    \centering
    \includegraphics[width=0.45 \textwidth]{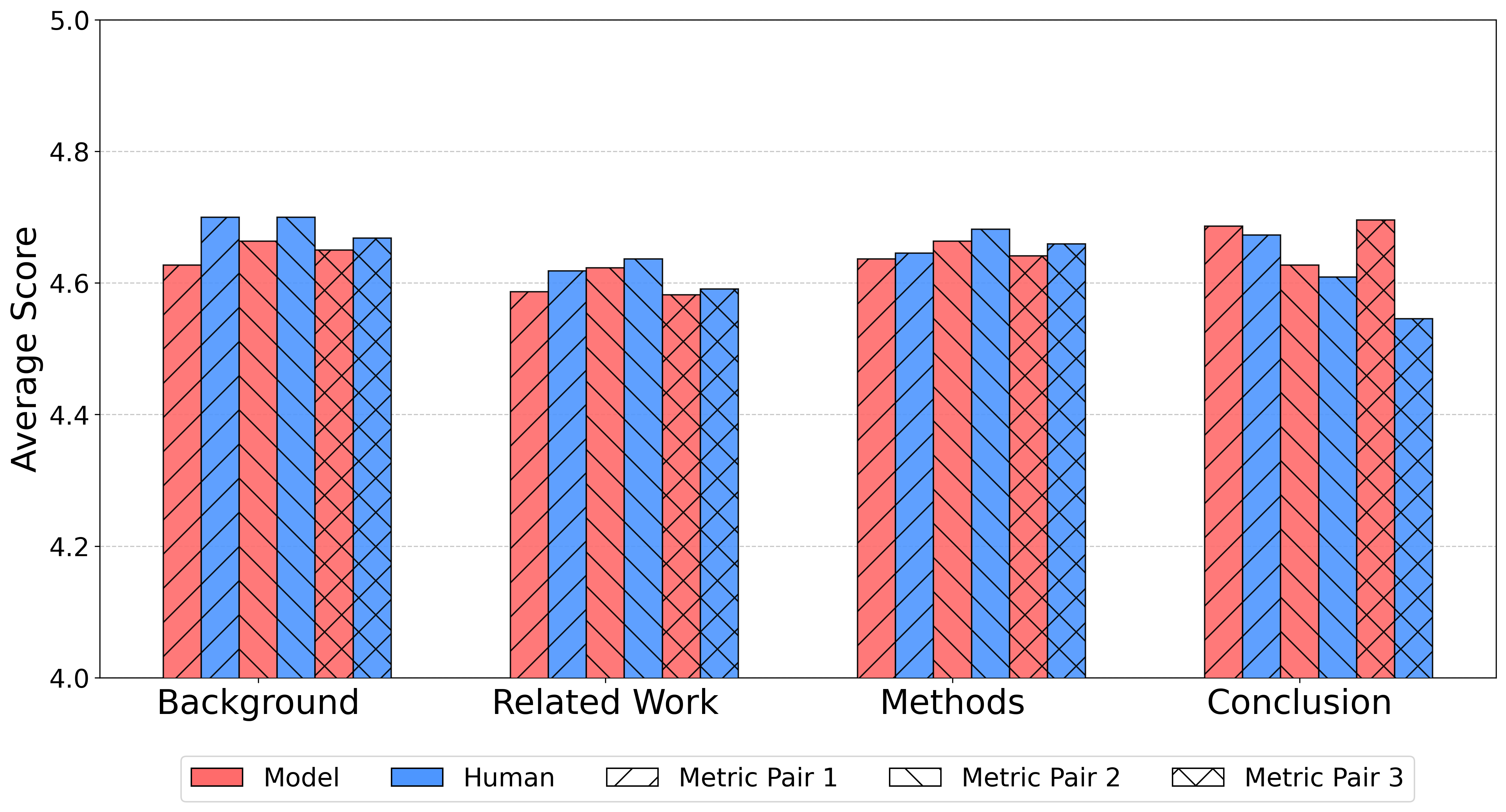}
    \caption{Human vs Model Writing Quality Comparison}
    \label{fig:human_model}
\end{figure}

\subsection{Impact of Training Dataset Scale on Model Performance}

\begin{figure}[!htbp]
    \centering
    \includegraphics[width=0.45 \textwidth]{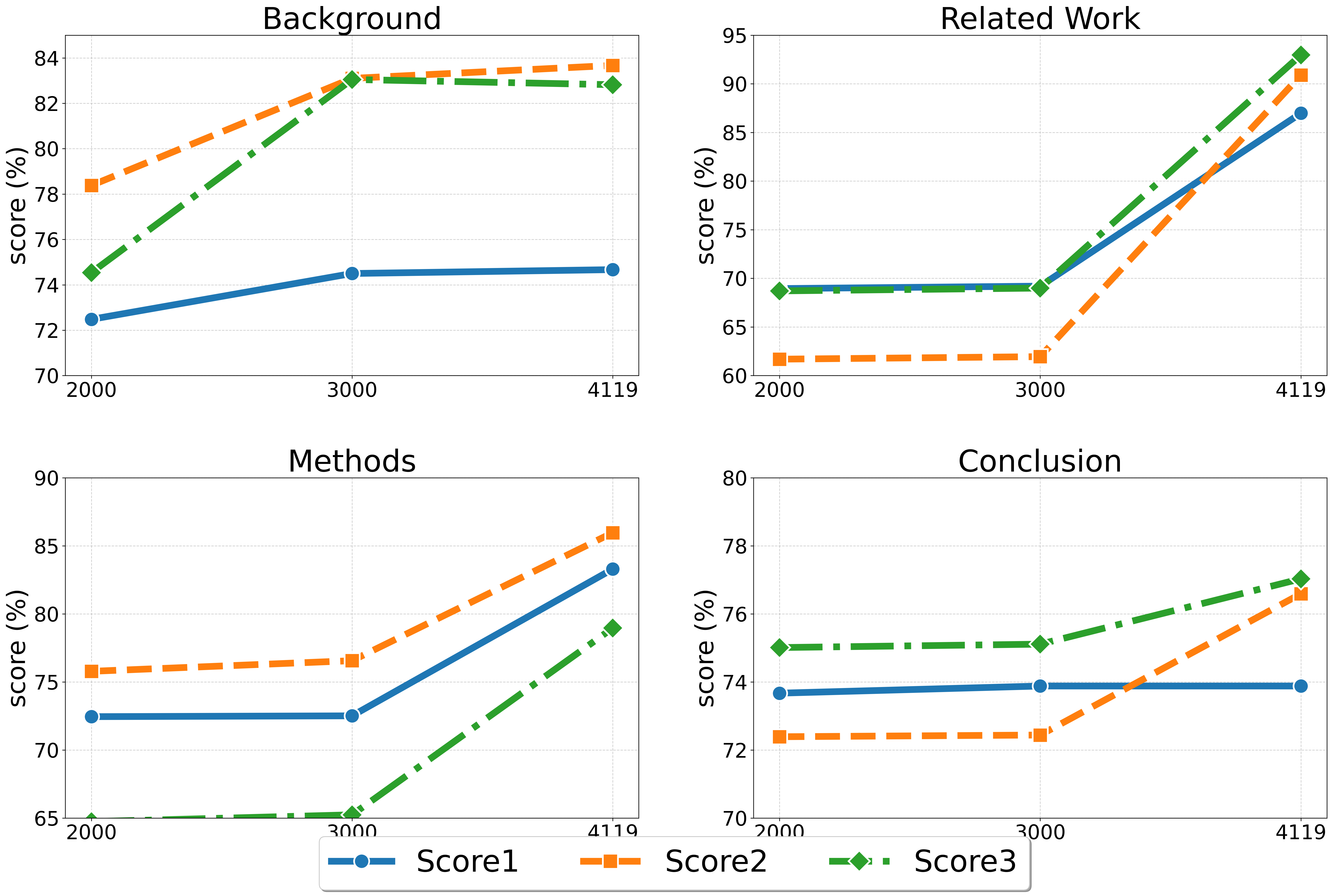}
    \caption{Impact of Training Sample Size on Multi-Dimensional Scoring Metrics}
    \label{fig:human_model}
\end{figure}

This experiment examines the relationship between training dataset size and model performance using three dataset scales: 2,000, 3,000, and 4,119 samples. As shown in Figure \ref{fig:human_model}, model scores exhibit a non-negative correlation with dataset size, increasing or remaining stable as the dataset expands. This trend is attributed to the model's enhanced ability to generate more accurate reports, as larger datasets provide a broader range of patterns and linguistic structures. During inference, the model is more likely to identify highly similar reference reports, leading to improved performance with larger training datasets.

\section{Conclusion}\label{sec:Conclusion}
Our Feedback-Refined Agent Methodology (\textbf{FRAME}) demonstrates significant improvements in medical paper generation, with DeepSeek V3 and GPT-4o Mini showing average performance gains of $9.91\%$ across $40$ evaluation dimensions, while Qwen 1.5 32B achieves a $1.5\%$ improvement. Human evaluations reveal that \textbf{FRAME}-generated papers achieve comparable quality to human-authored works ($92.80\%$ vs $92.88\%$), particularly excelling in conclusion synthesis. The proposed tripartite training architecture and structured dataset construction method effectively address key challenges in medical research automation, though limitations in retrieval dependency and offline dataset usage suggest potential for future enhancements through adaptive Generator-Evaluator-Reflector retrieval strategies and dynamic learning from external sources.

\section*{Limitations}
While the method paradigm brings significant advancements and proves effective, it is also subject to certain limitations that merit discussion. 

\textbf{Retrieval Dependency} Our study focused exclusively on a single retrieval step conducted prior to each section's generation. However, suboptimal retrieval quality may indirectly compromise the performance of our core modules. To address this, we propose that future work explore an adaptive, multi-round retrieval strategy that dynamically interacts with these core components, offering a promising direction for enhancing overall effectiveness.

\textbf{Offline Dataset} We build our training dataset by downloading the paper from the website. This offline method could limit our method engage with the latest paper. A promising avenue for further research lies in developing models that actively learn to search the information from the external search engine to improve information retrieval.

\textbf{Experimental Constraints} As an auxiliary tool for scientific writing, our method is designed to assist researchers in rapidly drafting papers based on existing topics and experimental results. However, due to the inherently specialized nature of medical research, our Agent cannot directly assist in conducting experiments or verifying the accuracy of experimental outcomes. Consequently, all experiments and analyses in this study are predicated on the assumption that the underlying papers do not contain fabricated or flawed data. The use of incorrect or incomplete data could severely mislead the paper generation process, potentially compromising the reliability of the output. Future improvements should include mechanisms to validate experimental data integrity or incorporate human-in-the-loop verification for critical research components.

\section{Acknowledge}
This work was supported in part by the Guangdong Provincial Key Laboratory of Human Digital Twin (Grant No. 2022B1212010004), the Guangzhou Basic Research Program (Grant No. SL2023A04J00930), and the Shenzhen Holdfound Foundation Endowed Professorship. Additional support was provided by the Anhui Provincial Major Science and Technology Project (Grant Nos. 202303a07020006 and 202304a05020071), the Anhui Provincial Clinical Medical Research Transformation Project (Grant No. 202204295107020004), and the National Key Research and Development Program of China Engineering Science and Comprehensive Interdisciplinary Special Project (Grant No. 2024YFF0507603).

\bibliography{acl_latex}

\clearpage
\onecolumn  
\appendix
\section{Ethical Considerations}
All personnel involved in the evaluation process participated voluntarily and received ample compensation. All data used in our experiment is sourced from arXiv and is allowed for non-commercial use. The core sections of the paper and all experiments were completed by humans, with AI only assisting in polishing the language and wording.

\section{Prompt used in FRAME}
Due to the space limit, we released our prompt used during the agent train stage. The remaining details will be available in the code.

\makeideabox{Extractor Prompt (Taking Conclusion as an example)}{
\textbf{Role:} Paper Analyst

\textbf{Task:} Identify and analyze the real-world problems addressed by the provided article paragraph.
    
\textbf{Requirements:}
\begin{itemize}
    \item If the input only contains the paragraph, then extract the conclusion based on the paragraph. If the input includes historical records from previous instances, it is necessary to reference both the paragraph and the previous scores and reasons to provide a better conclusion.
    \item Output format: Describe the analysis results in natural language and output in JSON format, which should only contain a single key-value pair with the key "conclusion" and the value being the conclusion description.
\end{itemize}

\textbf{Input Role:}
\begin{itemize}
    \item Current content: The article paragraph.
    \item Previous evaluations: Historical records or scores from previous evaluations.
\end{itemize}
}

\makeideabox{Checker Prompt (Taking Conclusion as an example)}{
\textbf{Role:} Conclusion Evaluator

\textbf{Task:} Evaluate the conclusion extracted from the provided article paragraph.

\textbf{Requirements:}
\begin{itemize}
    \item Input format: The input should consist of two parts: the article paragraph and the extracted conclusion.
    \item Output format: Return a JSON object containing two key-value pairs:
    \begin{itemize}
        \item ``score'': A numerical score from 0 to 10 (incremented by 0.1) indicating the relevance of the conclusion to the article paragraph.
        \begin{itemize}
            \item 0-2: Completely irrelevant
            \item 2-4: Mostly irrelevant
            \item 4-6: Partially relevant
            \item 6-8: Mostly relevant
            \item 8-10: Highly relevant
        \end{itemize}
        \item ``reason'': A textual explanation for the assigned score.
    \end{itemize}
\end{itemize}
}

\makeideabox{Generator Prompt (Taking Conclusion as an example)}{
\textbf{Task:} Based on the provided information from multiple related papers, we kindly request you to synthesize a comprehensive and well-structured conclusion section for a new research paper addressing this specific topic, ensuring that it integrates key findings and implications while maintaining academic rigor.

\textbf{Input:}
\begin{itemize}
    \item Research question: [question]
    \item Background: [background]
    \item Related Work: [related works]
    \item Methods: [methods]
    \item Result: [result]
\end{itemize}

\textbf{Output Format:} Please return your response in the following JSON format.

\textbf{Requirements:}
\begin{itemize}
    \item Summarize the key findings and their significance 
    \item Connect back to the research question
    \item Discuss implications and potential impact
    \item Acknowledge any limitations
    \item Suggest future research directions
\end{itemize}
}

\makeideabox{Reflector Prompt (Taking Conclusion as an example)}{
Please analyze the following conclusion section generation and provide improvement suggestions:

Topic: [topic]
    
1. Model Prediction (Generated Conclusion):
[prediction]
    
2. True Answer (Reference Conclusion):
[reference]
    
3. Evaluator's Comments:
[evaluator comments]
    
4. Evaluation Scores:
\begin{itemize}
    \item Comprehensiveness: [comprehensiveness score]/5.0
    \item Impact: [impact score]/5.0
    \item Future Direction: [future direction score]/5.0
\end{itemize}
    
Please analyze the gaps between the generated content and the reference, considering the evaluator's feedback and scores.
Focus on identifying specific areas for improvement and providing actionable suggestions.
    
THOUGHT:
Analyze the following aspects:
\begin{itemize}
    \item Results Summary:
    \begin{itemize}
        \item What key findings need better emphasis?
        \item Which results need clearer interpretation?
        \item How can the significance be better conveyed?
    \end{itemize}
        
    \item Research Impact:
    \begin{itemize}
        \item What theoretical contributions need better articulation?
        \item Which practical implications should be highlighted?
        \item How can the broader impact be better demonstrated?
    \end{itemize}
        
    \item Future Directions:
    \begin{itemize}
        \item What promising research directions should be added?
        \item How can future work suggestions be more specific?
        \item Which limitations need better connection to future work?
    \end{itemize}
        
    \item Integration and Synthesis:
    \begin{itemize}
        \item How can findings be better connected to research goals?
        \item What overarching themes need better emphasis?
        \item How can the conclusion be more compelling?
    \end{itemize}
\end{itemize}
}

\section{Specific examples of comments generated by the reflector}
\makeideabox{Background Section Generation Report}{
\textbf{Evaluation Results:}

\textbf{Scores:}
\begin{itemize}
    \item Completeness: 1.75/5
    \item Relevance: 2.0/5
    \item Organization: 2.0/5
\end{itemize}

\textbf{Evaluator Comments:}
\begin{itemize}
    \item \textbf{Content Coverage Comparison:}
    \begin{itemize}
        \item The depth and breadth of coverage in the content to evaluate are adequate for its specific topic but lack the detailed methodological and application context provided in the reference.
        \item (Some comments have been omitted due to length constraints.)
    \end{itemize}
    
    \item \textbf{Completeness Score and Rationale:}
    Based on the above analysis, the content receives a Completeness score of 1.75 because it covers the main concepts related to thymic involution and hyperplasia quantification but lacks detailed methodological discussion and specific application examples as seen in the reference.
    
    \item \textbf{Relevance Alignment:}
    \begin{itemize}
        \item The content does not deviate from its research focus but does not expand its relevance to broader societal impacts as effectively as the reference.
        \item (Some comments have been omitted due to length constraints.)
    \end{itemize}
    
    \item \textbf{Relevance Score and Rationale:}
    The content scores 2.0 in Relevance for its strong alignment with the research question but lacks the broader societal impact demonstrated in the reference.
    
    \item \textbf{Organizational Comparison:}
    \begin{itemize}
        \item The organization falls short of the reference's clarity and balance in presenting a detailed methodological approach and its validation.
        \item (Some comments have been omitted due to length constraints.)
    \end{itemize}
    
    \item \textbf{Organization Score and Rationale:}
    The content scores 2.0 in Organization for its logical flow and structure but falls short in methodological depth and balance compared to the reference.
\end{itemize}

\textbf{Overall Gap Analysis:}
\begin{itemize}
    \item The key qualitative differences from the reference include the lack of detailed methodological discussion, specific application examples, and broader societal relevance.
    \item (Some comments have been omitted due to length constraints.)
\end{itemize}

\textbf{Reflection Analysis:}
The content effectively outlines the thymus's immune function and CT imaging challenges but lacks methodological depth, specific applications, and societal relevance. While well-organized, it would benefit from clearer transitions and a more detailed methodological progression.

\textbf{Improvement Suggestions:}
\begin{itemize}
    \item \textbf{Content Improvements:}
    \begin{itemize}
        \item Include detailed methodological discussion, such as the specific image processing and machine learning techniques used for thymic quantification.
        \item (Some comments have been omitted due to length constraints.)
    \end{itemize}
    
    \item \textbf{Relevance Improvements:}
    \begin{itemize}
        \item Ensure all content is directly aligned with the research question by removing or revising any information that does not contribute to the understanding of automatic thymic quantification.
        \item (Some comments have been omitted due to length constraints.)
    \end{itemize}
    
    \item \textbf{Structure Improvements:}
    \begin{itemize}
        \item Add clearer transitions between topics to guide the reader through the methodological development and its validation.
        \item (Some comments have been omitted due to length constraints.)
    \end{itemize}
    
    \item \textbf{Style Improvements:}
    \begin{itemize}
        \item Make the explanation clearer by defining technical terms and providing examples where necessary.
        \item (Some comments have been omitted due to length constraints.)
    \end{itemize}
\end{itemize}
}

\end{document}